# Edge-Cloud Collaborated Object Detection via Difficult-Case Discriminator

Zhiqiang cao, Zhijun Li, Pan Heng, Yongrui Chen, Daqi xie, Liu jie

*Abstract*—As one of the basic tasks of computer vision, object detection has been widely used in many intelligent applications. However, object detection algorithms are usually heavyweight in computation, hindering their implementations on resource-constrained edge devices. Current edge-cloud collaboration methods, such as CNN partition over Edge-cloud devices, are not suitable for object detection since the huge data size of the intermediate results will introduce extravagant communication costs. To address this challenge, we propose a small-big model framework that deploys a big model in the cloud and a small model on the edge devices. Upon receiving data, the edge device operates a difficult-case discriminator to classify the images into easy cases and difficult cases according to the specific semantics of the images. The easy cases will be processed locally at the edge, and the difficult cases will be uploaded to the cloud. Experimental results on the VOC, COCO, HELMET datasets using two different object detection algorithms demonstrate that the small-big model system can detect 94.01%-97.84% of objects with only about 50% images uploaded to the cloud when using SSD. In addition, the small-big model averagely reaches 91.22%-92.52% end-to-end mAP of the scheme that uploading all images to the cloud.

*Index Terms*—Object detection; edge-cloud collaboration; neural networks; small-big model; difficult-case discriminator.

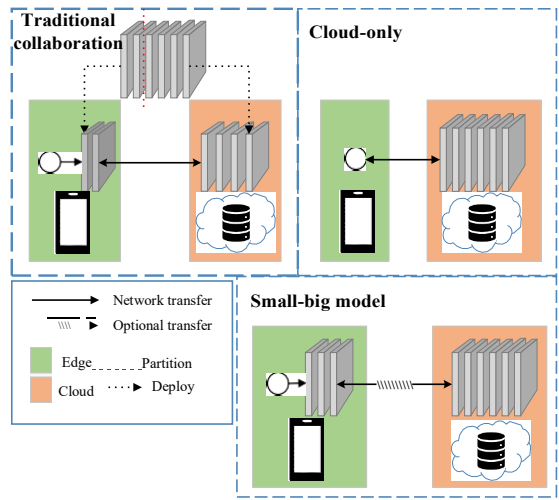

Fig. 1. Existing methods and small-big model system.

## I. INTRODUCTION

Convolutional Neural Networks (CNNs) [39] have been evolved into the most commonly adopted machine learning techniques. Due to their superior performances [21], [30], [34]–[37], people have witnessed their successful applications in a broad spectrum of domains from computer vision to natural language processing.

However, CNN-based applications typically require a tremendous amount of computation that impedes their wide adoption in resource-constrained mobile devices. Generally, due to the excessive resource demand of CNNs, the end devices have to transmit the data to a cloud platform, where task-specific deep neural network (DNN) models are executed to analyze the data, and the results are transmitted back to the edge devices (Fig. 1-top right) [22]–[25]. However, within such a cloud-centric framework, large amounts of data (e.g., images and videos) have to be uploaded to the remote cloud via a wide-area network, resulting in high end-to-end latency and large bandwidth requirement.

To break this limit, the collaboration between devices and cloud has been proposed, which makes use of both the high computation ability of the cloud and the local processing ability of the edge (Fig. 1-top left) [6], [26], [27], [45]. Recent works divide a model into two parts, one running in the cloud and the other running at the edge. The division is based on some measurable indicators, such as the waiting time of each layer of the model, the size of the output data, the bandwidth constraints, and so on. At run time, the edge device executes the first part of the model and transmits the intermediate results to the cloud. The cloud continues the model execution and returns the final result to the device. Overall, these approaches allow tuning the partition of CNN that will be executed on each platform based on their capabilities.

Although existing Edge-Cloud collaborated approaches can be easily applied in image classification, they are not suitable for object detection. The main reason is, even if the model can be partitioned for object detection, the intermediate results will contain a lot of features. This will introduce a lot of data that has to be transmitted from the edge to the cloud, even more than the image itself, thus incurring extra bandwidth consumption and inference delay.

To address this issue, we propose the small-big model framework to realize Edge-Cloud collaboration in object detection (Fig. 1-bottom). This framework is to deploy a lightweight object detection model at the edge and a heavyweight model in the cloud.

A difficult-case discriminator is operated at the edge to categorize images into "easy cases" and "difficult cases", according to their specific semantics extracted from the predictions of the model deployed on the edge device. Images classified as difficult cases will be uploaded to the cloud and processed by a heavyweight model. And images classified as easy cases will be processed locally at the edge by a lightweight model. Undoubtedly, this framework can efficiently exploit the computing resources of the edge devices, save the communication cost between the edge and the cloud, and effectively reduce the average inference latency.

Overall, our work makes the following key contributions:
- We propose a novel CNN model deployment framework for edge-cloud collaboration. Existing edge-cloud collaboration solutions typically treat CNN as a computation graph and partition it between device and cloud, while our framework deploys a small model at the edge and a big model in the cloud, and transmits data optionally.
- To deploy edge-cloud collaboration, we address several challenges, including: i) How to define difficult cases. ii) How to design a difficult-case discriminator which can accurately predict whether the image can be processed by the on-device model. iii) How to design a lightweight small model which can achieve a flexible trade-off between computation resources and overall performances.
- We evaluate our framework on two object detection algorithms, three small models, three datasets, and a real-world application. The experiment results show that on average our framework saves communication cost by 50%, reduces 32% inference time, and the mean Average Precision (mAP) can reach 91.22%-92.52% that of all images are uploaded to the cloud. It is proved that this framework is generic and can be applied to various types of one-stage object detection algorithm to reduce communication cost and latency by edge computing.

## II. RELATED WORKS AND MOTIVATION

Related efforts can be classified into three categories: deep learning on resource-constrained devices, cloud offloading and partitioned execution.

### A. Model Compilation for Edge Devices

Recently, there are many interests on optimizing deep learning models for embedded MCUs [10]–[13]. Some works develop appropriate models targeting specific resource-constrained edge devices such as IoT devices. The most mainstream is still model compression technology [28], [29], [33], [38]. Many smartphone-class device companies like Google and Amazon offer model compression tools to make it easy for developers to optimize deep learning models for mobile devices. The two lightweight models adopted in this paper, mobile net v1 and v2 [1], [14] are from Google. For many applications (e.g. automatic driving), high precision is required for object detection algorithms. However, at present, there is no method suitable for object detection that can achieve automatic compression without losing considerable accuracy.

### B. Cloud Offloading

This is a traditional method for edge devices, and there have been more than a decade of works on optimizing cloud offload [2]–[5], [15]. They primarily deal with variations in network delay and bandwidth at the network layer. The advantage of this method is a high accuracy, and the disadvantage is the large delay and waste of the computing power at the edge.

### C. Model Partition

Several recent efforts have investigated on the partition of deep learning models across cloud and edge. One of the most prominent works, Neurosurgeon [6], partitions CNN between a device-mapped head and a cloud-mapped tail, and selects a single split point based on the device and cloud load as well as the network conditions. There have been some works that extend the idea of model partition [7]–[9], [17]. Edgent [16] proposed a method that merges cloud offloading with multi-exit models. In general, model partitioning needs to transmit intermediate results from the edge to the cloud, but the amount of intermediate data for object detection is quite large, even larger than the image itself, so this method is not suitable for object detection.

This paper aims at addressing this issue, i.e., how to deploy object detection models on both the cloud and the edge while taking into account latency, accuracy, and resource consumption of the whole system.

## III. SYSTEM DESIGN OVERVIEW

The small-big model system is mainly composed of three modules: small model, big model, and difficult-case discriminator. The small model and the difficult-case discriminator are deployed at the edge, and the big model is deployed in the cloud. The core part is the difficult-case discriminator, which is responsible for deciding whether an image can be locally processed (we call it an easy case), or has to be transmitted to the cloud (we call it a difficult case). Easy cases are processed at the edge-side lightweight model, and difficult cases are uploaded to the cloud. The decision is based on the inherent semantics of the image.

The main workflow of small-big model is illustrated in Fig. 2 and described as follows. Step *1*, the edge device captures the image, and inputs it into the small model for preliminary recognition. Step *2*, the preliminary result of small model detection will be input into the difficult-case discriminator. Step *3*, the difficulty of the image (i.e., easy-case or difficult-case) will be output and returned back to the small model. Step *4*, if it is labeled by the discriminator as a difficult case, the edge will upload the image to the cloud and hand it over to the big model for more accurately object detection. Step *5*, after processing the image, the output of the big model will be returned to the edge device as the final result. Step *6*, if the image is deemed as an easy case by the difficult-case discriminator, the detection result of the small model will be output as the final result. In general, the inference flow for an easy case is *1-2-3-6* and the inference flow for a difficult case is *1-2-3-4-5-6*.

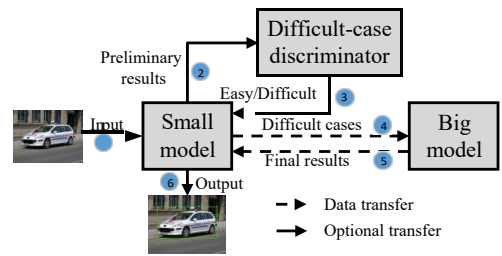

Fig. 2. The workflow of the small-big model system.

## IV. EDGE-CLOUD COLLABORATED OBJECT DETECTION MODEL

### A. Big Model

The current mainstream object detection algorithms are mainly based on deep learning models, which can be roughly divided into two categories: One-Stage and Two-Stage object detection algorithms. Two-Stage object detection algorithm, such as R-CNN [20] and Faster R-CNN [31], divides the detection problem into two stages. In the first stage, it generates candidate areas (i.e., Region Proposals), which contain the approximate location information of the objects. And in the second stage, it classifies the objects in the candidate areas and then refines their locations. On the other hand, One-Stage object detection algorithms, such as YOLO [19], SSD [18] and Corner Net [32], do not require the Region Proposal stage, and can directly generate categorical distributions and position coordinates of objects through Stage Value. Two-Stage algorithms perform better in accuracy, while One-Stage algorithms are generally much faster. Considering that Edge-Cloud collaboration focuses more on timeliness (e.g., object detection for video stream), we choose One-Stage algorithms as our large models[1]. Our design is suitable for any one-stage object detection algorithm, and we use two representative algorithms–YOLOv4 [40] and SSD in this paper for demonstration.

### B. Small Model

An object detector (one-stage) is typically composed of four parts: (i). Input, e.g., Image, Patches, Image Pyramid; (ii). Backbones/Base Network, e.g., VGG16 [21], ResNet-50 [37], CSPDarknet53 [41]; (iii). Neck, e.g., SPP [42], FPN [43], PAN [44]; (iv). Heads (dense prediction, anchor based), e.g., RPN [31], SSD, YOLO. The small model is not limited to a specific algorithm. Instead, it can be designed according to the user's needs. The main idea is to reduce the Neck and Backbones according to the structure of the big model, to reduce the amount of calculation. But it will inevitably weaken the recognition ability of images with some specific semantics.

For better demonstration, we choose SSD as the big model and build a small model as an example to elaborate on it. The base network of the small model is composed of VGG-Lite and Conv6&7. The VGG-Lite part is a cut-down version of VGG16 (similar to the deletion of VGG11), and then Conv6&7 is connected behind. After that, 8 layers of extra feature layers (Neck) are connected one by one. The detection process is also to imitate the SSD. Compared with SSD, the feature layer of 38*38 is removed. Additionally, different from VGG16, VGG-Lite removes 9 convolution layers and two pooling layers. Con6&7 is to adjust the scale of the feature layers to facilitate connection with the following extra feature layers. The structure of VGG-Lite + Conv6&7 is shown in Fig. 3.

[1]Our framework can also be applied for Two-Stage algorithms.

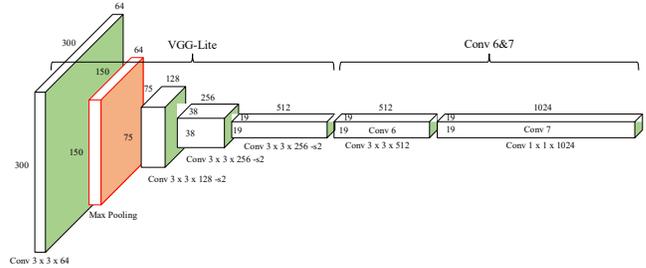

Fig. 3. The base network of the small model.

In SSD, the large-size feature maps are used to analyze small objects, while the small-size feature maps are used to analyze large objects. Therefore, we discard the feature map of 38*38 that will weaken the model's ability to detect small objects. On the other hand, the feature map of 38*38 provides 5776 default boxes (8,732 boxes in total). After discarding it, the small model loses 66% of default boxes. The reduction of a large number of default boxes will undoubtedly affect the recognition results of multi-object images and cause the missing of some objects. The method is the same for other object detection algorithms, i.e., replacing the lightweight base network, and then removing the large-size feature map.

In summary, these designs will largely reduce the amount of calculation and the size of the small model, making it adapt to the resource-constraint edge devices. On the other hand, compared with the big model, the small model is prone to missing objects with small object areas, as well as those in an image with a large number of objects.

## V. DIFFICULT-CASE DISCRIMINATOR

### A. What are Difficult Cases?

The difficult-case discriminator is the core part of the entire small-big model system, which controls the communication between the small model and the big model. Now the question is: what kind of images are difficult cases?

For image classification, the difficult cases can be discriminated by calculating the entropy value of the detection result, the compression ratio, and so on. However, object detection is much more complicated. For an image, the output of the object detection algorithm consists of dozens of bounding box predictions. Each object in the image will be marked in a bounding box, and the prediction of the object's class along with the confidence score will be given. The higher the score, the more likely the detection is correct. However, some undetected objects may exist due to various reasons.

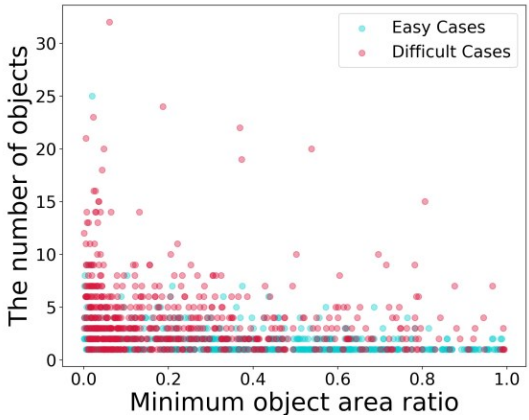

Fig. 4. The distribution of easy cases and difficult cases are based on two features-the number of objects and the minimum object area ratio.

Apparently, if there are undetected objects, the image should be uploaded to the cloud for further detection. Therefore, we define an image as a difficult case if the small model fails to detect all the objects in it and vice versa.

According to this definition, we can mark the dataset. The detection result of the big model is compared with the result of the small model. When the difference in the number of detected objects is greater than or equal to 1, it means that there is at least one undetected object. We will mark the image as a difficult case, and the opposite is an easy case. The confidence score threshold of the recognition boxes is 0.5, only the recognition boxes with a score value greater than 0.5 are considered as correctly identified objects.

### B. How to Discriminate Difficult Cases?

How to discriminate difficult cases can be regarded as a binary classification problem. First of all, the difficult-case discriminator is deployed on edge devices, it has to be lightweight. An intuitive way is to extract the output results of the image as features, and input them into a few Conv layers. However, we found that this method does not work well. The reason is, "difficult" and "easy" are very abstract concepts, it is unrealistic to bridge the semantic gap between the input features and the high-level semantics just by a few Conv layers.

Therefore, if a predictive model is directly built on the underlying features, it may be too complex to be implemented on edge devices. But if we can extract some good representations which reflect the high-level semantic features to some extent, we may build machine learning models in a relatively easy way.

As described in Sec.IV.B, when designing the small model, we found that the ability of the small model to recognize images with multiple objects and with small objects is weakened, due to the reduction of convolution layers and large-size feature maps. So we design the difficult-case discriminator based on the two inherent characteristics – the number of objects and the area ratio of objects.

Supposing that there is at least one missed object, then the object with the smallest target area ratio (i.e., the proportion of the target area to the entire image area) will be the most likely to be missed. So, we choose the number of objects and the proportion of the smallest object area as two features to assess whether an image is "difficult" or "easy". In a word, we exploit empirical knowledge to select good features. This can improve the performance of the difficult-case discriminator without complex network architectures.

To verify our conjecture, we analyzed these two features through Fig. 4. First, we trained the small model and the big model (SSD) on VOC07+12, i.e. VOC2007 trainval + VOC2012 trainval. Second, we calculated the number of objects as well as the minimum object area ratio for each image from its annotation. Third, via analyzing the detection results given by both models, we labeled each image as a difficult case or an easy case according to the aforementioned criterion. In the end, we use a red dot to represent a difficult case, a blue dot to represent an easy case to draw a scatter plot (Fig. 4). The figure clearly shows that difficult cases are concentrated in areas with a large number of objects or a small minimum object area ratio (i.e., the left and upper parts in the figure) while easy cases have fewer objects and a larger minimum object area ratio.

### C. How Discriminator Works?

For the purpose of distinguishing difficult cases from easy cases, we propose a simple and efficient method via a threshold model. The entire workflow is shown in Fig. 5. The input of the discriminator is the preliminary detection result of the image, given by the small model. And the output is the type of the image, namely the image is a difficult case or an easy case. Specifically, the workflow can be spitted into two major parts: one is to assess the object number along with the minimum object area ratio by processing the preliminary result, the other is to decide the type of the image (i.e., difficult or easy) on the basis of the assessed values and the preliminary detection result. Next, we will further explicate two parts in the discriminator's workflow respectively.

*1) **Predict**:* Via scrutinizing the detection results given by the small model, we find out that there is a notable gap between confidence scores of different object classes in a bounding box. That is, the confidence score of the class of the existent object is significantly higher than other classes.

For instance, as illustrated in Fig. 6, a bounding box consists of 5 elements – confidence score, x min, y min, x max, y max. The image has two objects in it, and, considering that any bounding boxes with confidence score lower than 0.5 would be ignored, the person (with 0.9818 score) can be detected while the dog, having no valid bounding box, cannot. However, it is worth noting that the confidence score of the dog, though being missed, is 0.2507, remarkably higher than any other object class that does not really exist, like 'cat' (0.0735) or 'bottle' (0.0572).

Inspired by such observation, we obtain a threshold for filtering out noise boxes through regression, which will be discussed later in detail. Any bounding box scored lower than the threshold would be considered as a noise box and be neglected. After removing all the noise boxes, we can calculate the estimated number of objects and the estimated minimum object area ratio in the image.

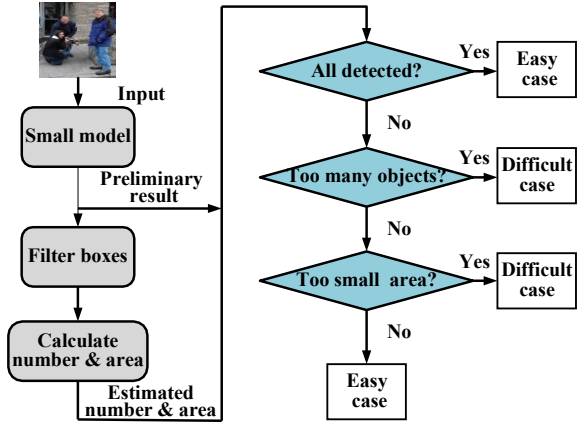

Fig. 5. The workflow of the difficult-case discriminator.

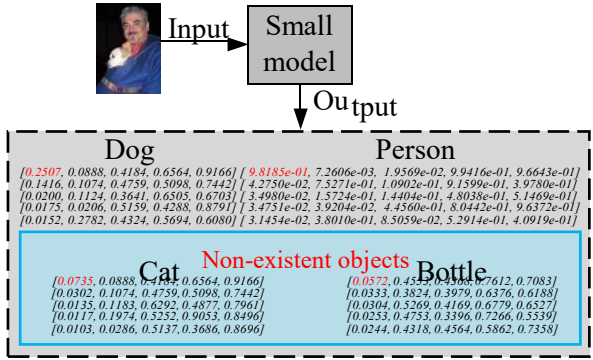

Fig. 6. The preliminary inference result of a single image. The person is successfully recognized, while the dog is missed.

Moreover, in the first step of the workflow of our discriminator, we will compare the number of objects predicted by the small model with the number of bounding boxes after filtration, which serves as the estimated number of objects in the image. If equals, the discriminator will regard the image as an easy case. This simple step makes sense owing to the fact that, in essence, the predicting process of the small model is to filter out objects scored lower than the threshold, which in many occasions is 0.5, while the way to estimate the number of objects in the image is just the same as the prediction process, merely using different threshold, like 0.15-0.35. Therefore, if the number of predicted objects equals the estimated number of objects, which means that the value of the threshold does not make a difference and there is no uncertain object, the image is presumably an easy case.

*2) Discriminate*: On the basis of the estimated number of objects and the estimated minimum object area ratio in the image we expound in the previous section, as well as the preliminary detection result of the image given by the small model, we can come to the final decision on whether the image is a difficult case through the following three steps.

1. Compare the number of objects predicted by the small model with the estimated number of objects. If the two are equal, it means all objects in the image are very likely to be detected. Then it is classified as an easy case. Otherwise, continue the procedure.

2. Compare the estimated number of objects with the threshold for the number of objects. If the former is greater, then the image is regarded as a difficult case. Otherwise, continue the procedure.

3. Compare the estimated minimum object area ratio with the threshold for the minimum object area ratio. If the former is smaller, then the image is regarded as a difficult case. Otherwise, the image is an easy case.

*D. How to Set Optimal Thresholds?*

The whole workflow revolves around three thresholds, namely the threshold for confidence score, the threshold for the number of objects, and the threshold for minimum object area ratio. To fulfill our threshold model, we need to acquire the optimal values for the three thresholds from the training dataset. To get the optimal threshold for filtering out noise boxes via regression, we apply the following loss function.

$$L = N^{predict} - N^{truth} \quad (1)$$

In the above formula, $N^{predict}$ is the total number of objects predicted by the small model, and $N^{truth}$ is the total number of objects calculated from annotations in the dataset, i.e. ground truths. When the value of L reaches the bottom, the confidence score is the optimal value of the threshold.

Then, we can exploit regression to calculate the threshold for the number of objects and the threshold for the minimum object area ratio when the accuracy reaches the top. Please note that we input the true number of objects and minimum object area ratio into the discriminator here, instead of inputting the estimated values, because we do regression on the training dataset, which contains ground truths.

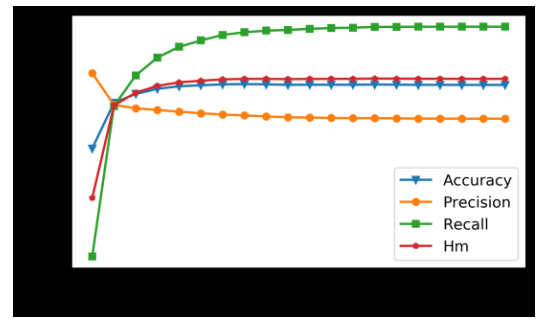

Fig. 7. The performances of discriminator when we fix the threshold of the number of object to 2 and change the threshold of the minimum object area ratio.

TABLE I
THE RESULTS OF THE DIFFICULT-CASE DISCRIMINATOR ON THE TRAINING DATASET AND THE TEST DATASET.

|  | Accuracy | F1 | Precision | Recall |
|---|---|---|---|---|
| Ground Truth | 85.35% | 0.8665 | 77.51% | 98.24% |
| Predicted | 78.35% | 0.7732 | 78.38% | 76.29% |

TABLE II
THE MODEL SIZE AND NUMBER OF COMPUTING OPERATIONS OF THREE SMALL MODELS.

|  | Model size(MB) | Pruned(%) | FLOPs(Billion) |
|---|---|---|---|
| Small model 1 | 18.50 | 81.55 | 5.60 |
| Small model 2 | 11.55 | 88.48 | 5.31 |
| Small model 3 | 6.50 | 93.52 | 1.31 |
| SSD | 100.28 | - | 61.19 |

TABLE III
THE MAP WHEN USING SMALL MODEL 1.

|  | Big model mAP(%) | Small model mAP(%) | End-to-end mAP(%) | Upload ratio (%) |
|---|---|---|---|---|
| 07 | 70.76 | 41.28 | 62.68 | 51.47 |
| 07+12 | 77.41 | 51.34 | 71.61 | 51.23 |
| 07++12 | 72.31 | 49.02 | 66.42 | 50.76 |
| COCO | 42.18 | 27.78 | 38.76 | 52.09 |
| Average | - | - | - | 51.32 |

TABLE IV
THE NUMBER OF DETECTED OBJECTS WHEN USING SMALL MODEL 1.

|  | Big model | Small model | End-to-end | End-to-end/Big model(%) |
|---|---|---|---|---|
| 07 | 9055 | 4759 | 8325 | 93.00 |
| 07+12 | 9628 | 5511 | 9100 | 94.51 |
| 07++12 | 8434 | 5202 | 7852 | 95.07 |
| COCO | 7996 | 4353 | 7424 | 92.84 |
| Average | - | - | - | 94.01 |

Based on the experiments, we find out that the optimal threshold for the number of objects is 2 and the threshold for the minimum object area ratio is 0.31. As shown in Fig. 7, accuracy (difficult cases are viewed as positive examples in this figure.) reaches 85.35%, recall reaches 98.24%, precision reaches 77.51%, hm (harmonic mean) reaches 0.8665. After obtaining three thresholds, we can distinguish the test dataset. The results are shown in Table I, accuracy reaches 78.35%.

## VI. EVALUATION

### A. Experiment Setup

One set of experiments are conducted in a computer with AMD Ryzen9 5900HX CPU, NVIDIA RTX3060 GPU, and 32G RAM. Another set of experiments are conducted in an NVIDIA Jetson Nano client and a server, which is the same computer used in the previous experiments. The client and the server are connected via WLAN.

*1) Datasets:* three datasets are used in our experiments:

**07**: PASCAL VOC2007, train: VOC2007 trainval (5011 images), test: VOC2007 test (4952 images).

**07+12**: PASCAL VOC2007 + VOC2012, train: union of VOC2007 trainval (5011 images) and VOC2012 trainval (11540 images), test: VOC2007 test (4952 images).

**07++12**: PASCAL VOC2007 + VOC2012, train: union of VOC2007 trainval + test (9963 images) and the VOC2012 trainval (6588 images), test: VOC2012 trainval (4952 images, randomly chosen from VOC2012 trainval). 07+12 and 07++12 are cross-validation for each other.

**COCO**: COCO trainval 135k. MS COCO has as many as 80 classes of objects. Since our research focuses on edge-cloud collaboration rather than pure object detection, we select part of the objects. We select a total of 98,267 images containing 18 classes of objects, which are the same 18 classes as in the VOC dataset. 5% of them are randomly selected and used as the test set (4914 images), the remaining ones serve as the training set (93353 images).

**Helmet**: The dataset (helmet dataset) comes from the KubeEdge open source sub-project Sedna, which is derived from the images collected by the camera on building sites. The images come from a real scene, so there are various classes: blur, occlusion, water stains, smoke, insufficient light, etc., which can test the robustness of our edge-cloud collaborative object detection model.

*2) Metrics:* We use the following six metrics to measure our algorithm.

1. **mAP (mean Average Precision)**: This metric reflects the accuracy of the object detection algorithm.

2. **The number of detected objects**: This metric reflects whether our method can successfully detect objects as many as possible.

3. **Upload ratio**: It is defined as the ratio of the number of images uploaded to the cloud to the total number of images. This metric reflects the bandwidth consumption of the communication link from the edge to the cloud.

4. **Model size**: This metric reflects the storage resources occupied by the small model at the edge device.

5. **Pruned**: The reduction ratio of the small model relative to the large model.

6. **Number of computing operations (FLOPs)**: This metric reflects the computing power consumption of small models on the edge device, and directly affects whether the small-big model system can be successfully deployed in the real world.

### B. The Performance of the Small-Big Model system

In this part, we test the method on three different datasets, three different small models, and two different big models. The model size and FLOPs of three small models are shown in Table II. All the small models are lightweight models with pruned above 80%.

*1) The Small Model 1:* The small model 1 is designed by us and described in Sec. IV. The results are shown in Table III and Table IV. From the results we can see, using our framework, about 50% of images are uploaded to the cloud, and the end-to-end mAP of our small-big model system is 3.42%-8.08% lower than the cloud-only method. In other words, our method can save 50% of bandwidth consumptions and achieve 91.22% recognition precision compared with the method that all data are transmitted to the cloud.

TABLE V
THE MAP WHEN USING SMALL MODEL 2.

|  | Big model mAP(%) | Small model mAP(%) | End-to-end mAP(%) | Upload ratio (%) |
|---|---|---|---|---|
| 07 | 70.76 | 49.62 | 64.00 | 52.16 |
| 07+12 | 77.41 | 56.24 | 71.38 | 51.97 |
| 07++12 | 72.31 | 56.01 | 67.80 | 51.69 |
| COCO | 42.18 | 32.66 | 41.46 | 50.65 |
| Average | - | - | - | 51.61 |

TABLE VI
THE NUMBER OF DETECTED OBJECTS WHEN USING SMALL MODEL 2.

|  | Big model | Small model | End-to-end | End-to-end/ Big model(%) |
|---|---|---|---|---|
| 07 | 9055 | 6264 | 8810 | 97.29 |
| 07+12 | 9628 | 6486 | 9320 | 96.80 |
| 07++12 | 8434 | 6393 | 8323 | 98.68 |
| COCO | 7996 | 6257 | 7884 | 98.60 |
| Average | - | - | - | 97.84 |

TABLE VII
THE MAP WHEN USING SMALL MODEL 3.

|  | Big model mAP(%) | Small model mAP(%) | End-to-end mAP(%) | Upload ratio (%) |
|---|---|---|---|---|
| 07 | 70.76 | 42.00 | 64.29 | 51.99 |
| 07+12 | 77.41 | 48.47 | 72.24 | 51.85 |
| 07++12 | 72.31 | 44.84 | 66.42 | 51.99 |
| COCO | 42.18 | 26.85 | 38.50 | 48.96 |
| Average | - | - | - | 51.19 |

TABLE VIII
THE NUMBER OF DETECTED OBJECTS WHEN USING SMALL MODEL 3.

|  | Big model | Small model | End-to-end | End-to-end/ Big model(%) |
|---|---|---|---|---|
| 07 | 9055 | 4889 | 8647 | 95.49 |
| 07+12 | 9628 | 5242 | 9079 | 94.29 |
| 07++12 | 8434 | 4645 | 8101 | 96.05 |
| COCO | 7996 | 6388 | 7917 | 99.01 |
| Average | - | - | - | 96.23 |

The number of objects detected by this method is only 6% less than that of all data uploaded to the cloud. In other words, this method only needs about half of the bandwidth resources to detect about 94% of the objects compared with the method that all data are transmitted to the cloud.

*2) The Small Model 2 (MobileNet v1)*: We use Google MobileNet v1 as the base network, extra feature layers unchanged. The results are shown in Table V and Table VI.

*3) The Small Model 3 (MobileNet v2)*: MobileNet v2 is an improved version of Google's MobileNet v1, the model size and calculation amount are greatly reduced. We use MobileNet v2 as the base network, and the other settings of the experiment are the same as before. The results are shown in Table VII and Table VIII.

On the mAP of the small model, MobileNet v2 is down 5.81%-11.53% compared to v1. But on the end-to-end map, v2 is only 0.28%-2.96% lower than v1, this just proves the superiority of our method and can make up for the difference in performance between small models.

In general, our method can reach an average of 91.22%-92.52% compared to the case of all images uploaded to the cloud in end-to-end mAP, and misses about 2.16%-6% of the objects, but it can save half of the bandwidth resources. And some images(easy cases) directly outputting the detection results of the small model can reduce delay and also make full use of the computing power at the edge device.

The performance of this method does not completely depend on the performance of the small model, it can even make up for the performance difference between some small models. The end-to-end mAPs of the three small models under the same data set are slightly different. Users can flexibly design the structure of the small model (base network) according to the edge device.

The test results are stable on all three different datasets and three different small models, which proves that this method is robust in the condition of different application scenarios.

### C. Evaluations on YOLOV4

The previous experiment results are all based on SSD, now we will test the versatility of our method in other one-stage object detection algorithms. In this experiment, we select MobilenNet v1 as the base network, and reduce the large-scale feature map to build a small model, and then set the big model as YOLOv4. The results on 07 and 07++12 are shown in Table IX and Table X. Because of the improved performance of YOLOv4, the number of difficult cases is fewer, and a high end-to-end mAP can be achieved with only 20% of the upload cloud ratio. Since our method is based on the inherent semantics of images, it is suitable for all one-stage object detection algorithms.

TABLE IX
THE MAP OF WHEN USING YOLOv4.

|  | Small model mAP(%) | Big model mAP(%) | End-to-end mAP(%) | Upload ratio (%) |
|---|---|---|---|---|
| 07 | 73.64 | 83.48 | 79.52 | 20.90 |
| 07+12 | 79.72 | 90.02 | 85.78 | 21.32 |
| Average | - | - | - | 21.11 |

TABLE X
THE NUMBER OF DETECTED OBJECTS OF WHEN USING YOLOV4.

|  | Big model | Small model | End-to-end | End-to-end/ Big model(%) |
|---|---|---|---|---|
| 07 | 11098 | 10509 | 10985 | 98.98 |
| 07+12 | 11574 | 10478 | 11360 | 98.15 |
| Average | - | - | - | 98.57 |

TABLE XI
THE EXPERIMENTAL RESULTS ON HELMET UNDER REAL-WORLD CLOUD-EDGE COLLABORATION.

|  | Edge-only | cloud-only | Our method |
|---|---|---|---|
| mAP | 75.04 | 92.40 | 86.07 |
| Number of detected objects | 940 | 1135 | 1119 |
| Total inference time (s) | 47.13 | 264.76 | 179.79 |
| Upload ratio(%) | - | - | 51.19 |

### D. Evaluations on Real-World Edge-Cloud Devices and Real Dataset

On building sites, all workers must wear safety helmets. So the dataset is from some scenarios of some workers' working. We deploy small model 1 on the Jetson Nano and SSD on the server, edge-cloud devices are connected via WLAN. The experimental results are shown in Table XI. The end-to-end mAP of our method is 6.33% lower than the cloud-only method, but our method saves 32% of inference time and 50% of bandwidth resources. The number of objects detected by this method is only about 1.41% less than that all data are uploaded to the cloud. It is proved that our method can be applied in real-world applications.

### E. Comparison with other Difficult-case Discriminators

Our difficult-case discriminator is based on two inherent semantics of the image: the number of objects and the minimum object area ratio. We compare our method with three other baseline difficult-case discriminating strategies: randomly uploading, blurred images uploading, and uploading according to the top-1 confidence score. We adopt the small model 1 and SSD as the big model.

*1) Randomly upload images to the Cloud:* We randomly selected 50% of images uploaded to the big model of the cloud, and the remaining images are left in the small model.

TABLE XII
THE MAP OF THE METHOD RANDOMLY UPLOADING IMAGES TO THE CLOUD.

|  | End-to-end mAP randomly(%) | End-to-end mAP our method(%) |
|---|---|---|
| 07 | 56.64 | 62.68 |
| 07+12 | 64.06 | 71.61 |
| 07++12 | 60.87 | 66.42 |
| COCO | 34.82 | 38.76 |

TABLE XIII
THE NUMBER OF DETECTED OBJECTS OF THE METHOD RANDOMLY UPLOADING IMAGES TO THE CLOUD.

|  | End-to-end/ Big model(%) our method | End-to-end/ Big model(%) randomly | Upload ratio (%) |
|---|---|---|---|
| 07 | 93.00 | 74.83 | 51.47 |
| 07+12 | 94.51 | 77.07 | 51.23 |
| 07++12 | 95.07 | 78.69 | 50.76 |
| COCO | 92.84 | 75.06 | 52.09 |
| Average | 94.01 | 76.41 | 51.32 |

TABLE XIV
THE MAP OF THE METHOD UPLOADING BLURRED IMAGES TO THE CLOUD.

|  | End-to-end mAP blurred(%) | End-to-end mAP our method(%) |
|---|---|---|
| 07 | 57.30 | 62.68 |
| 07+12 | 65.22 | 71.61 |
| 07++12 | 60.05 | 66.42 |
| COCO | 35.26 | 38.76 |

TABLE XV
THE NUMBER OF DETECTED OBJECTS OF THE METHOD UPLOADING BLURRED IMAGES TO THE CLOUD.

|  | End-to-end/ Big model(%) our method | End-to-end/ Big model(%) blurred | Upload ratio (%) |
|---|---|---|---|
| 07 | 93.00 | 73.13 | 50.84 |
| 07+12 | 94.51 | 75.90 | 50.84 |
| 07++12 | 95.07 | 78.33 | 50.42 |
| COCO | 92.84 | 70.14 | 50.48 |
| Average | 94.01 | 74.38 | 50.64 |

The results are shown in Table XII and Table XIII. The upload ratio is about 50%, in the end-to-end mAP, our method is 3.94%-7.55% higher than random uploading, and has an average improvement of 8.87%; in the number of detected objects, our method is 17.6% higher than random uploading.

*2) Upload Blurred Images to the Cloud:* There are many ways to define the ambiguity of an image. We choose the Brenner gradient function to define the ambiguity. The Brenner gradient function is one of the simplest gradient evaluation functions. It calculates the square of gray level differences between two pixels. The function is defined as follows:

$$\sum_y \sum_x |f(x+2, y) - f(x, y)|^2 \qquad (2)$$

where f(x, y) is the gray value of the pixel (x, y). The larger the value of the function, the clearer the image. The results are shown in Table XIV and Table XV. The upload ratio is about 50%. For the end-to-end mAP, our method is 5.37%-7.71% higher than the blurred images uploading strategy, improved by 12.41%; in terms of the number of detected objects, our method is 19.63% higher that.

*3) Upload Method Based on the Top-1 Confidence Score:* The confidence scores of the bounding boxes represent how confident the object detection algorithm is in the classification result. The higher the confidence score, the more confident the model is in the detection result. Therefore, it can be used to evaluate whether the image is a difficult case or an easy case. Take the top-1 of the recognition boxes of each type of object in a single image, and then add a total of 20 confidence scores for 20 categories (VOC) and then take the average value. According to this value, sort the entire dataset from large to small, take the first 50% of images in the small model and the last 50% upload to the cloud.

TABLE XVI
THE MAP OF THE METHOD UPLOADING IMAGES TO THE CLOUD BASED ON TOP-1 CONFIDENCE SCORE.

|  | End-to-end mAP confidence score(%) | End-to-end mAP our method(%) |
|---|---|---|
| 07 | 57.30 | 62.68 |
| 07+12 | 65.22 | 71.61 |
| 07++12 | 60.05 | 66.42 |
| COCO | 35.26 | 38.76 |

TABLE XVII
THE NUMBER OF DETECTED OBJECTS OF THE METHOD UPLOADING
IMAGES TO THE CLOUD BASED ON TOP-1 CONFIDENCE SCORE.

|  | End-to-end/ Big model(%) our method | End-to-end/ Big model(%) confidence score | Upload ratio (%) |
|---|---|---|---|
| 07 | 93.00 | 73.13 | 50.84 |
| 07+12 | 94.51 | 75.90 | 50.84 |
| 07++12 | 95.07 | 78.33 | 50.42 |
| COCO | 92.84 | 70.14 | 50.48 |
| Average | 94.01 | 74.38 | 50.64 |

The results are shown in Table XVI and Table XVII. The upload ratio is about 50%, in the end-to-end mAP, our method is 3.5%-6.39% higher than the uploading strategy based on top-1 confidence score, and has an average improvement of 10.51%; regarding the number of detected objects, our method is 11.61% higher than that strategy. In terms of the small model mAP, our method is 4.38%-11.66% higher and has an average improvement of 17.82%. This method is far better than the other two baselines, but it is still much worse than our semantic-based uploading strategy.

### F. Performances under Different Upload Ratios

The upload to cloud ratio is directly related to performance and resource consumption. Figure. 8 shows the end-to-end mAP with different upload ratios when using small model 1. As the upload ratio increases, the slope of the mAP curve drops, similar to a parabola. When the upload ratio is 50%, mAP has reached about 90% of mAP that all images uploaded to the cloud (similar to a parabola turning point). The same trend is shown for the number of detected objects in Figure. 9.

The number of detected objects increases slowly with the increase of the upload cloud ratio. When the upload cloud ratio is 50%, the number of detected objects has reached more than 94% of the cloud-only method. Similar to the mAP, the 50% is a parabola turning point.

### VII. DISCUSSIONS AND FUTURE WORKS

**Automatic object detection model compression.** Current model compression methods for image classification will lead to serious loss of accuracy when applied to object detection. In order to ensure accuracy, the small model deployed at the edge is designed manually at present. In the future, we will design automatic object detection model compression, that is, the users only need to select the object detection models in the cloud, and then a lightweight object detection model suitable for given edge devices and the difficult-case discriminator can be automatically obtained.

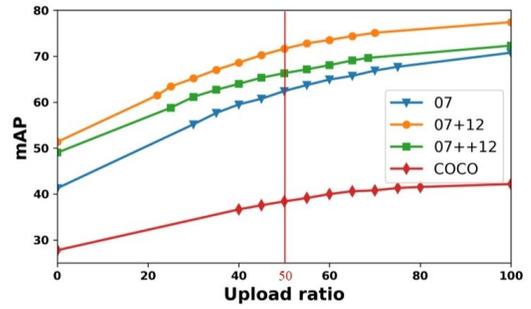

Fig. 8. The mAP (end-to-end) under different upload ratios.

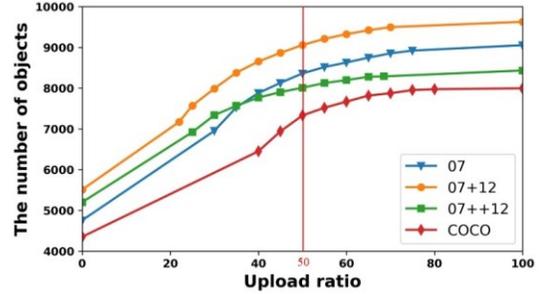

Fig. 9. The number of detected objects (end-to-end) under different upload ratios.

**Improving detection precision.** Compared with the method of uploading all images to the cloud, our method still loses about 8.78%-7.48% in the end-to-end mAP. Note that reinforcement learning has a significant effect on the neural architecture search of models for image classification, in the future, we will combine the semantic-based method with reinforcement learning to improve precision.

### VIII. CONCLUSION

In this paper, we investigate the problem of how to deploy the object detection algorithm in the edge-cloud collaboration context. From the object detection results on the edge device, we find that the images with missing objects are mostly with multi-objects or small objects. Inspired by this, we propose a novel framework that performs synergistic inference on both edge and cloud via the difficult-case discriminator. In this framework, a small model is deployed on the edge device, a big model is deployed in the cloud, and a difficult-case discriminator is designed to decide whether an image should be uploaded to the cloud. A set of techniques are presented for the difficult-case discriminator to achieve accurate prediction. Extensive experiments on the real-world dataset from Sedna and the real edge device demonstrate the effectiveness of the small-big model framework on various datasets and object detection algorithms. It is shown that the small-big model framework reaches an average 91.22%-92.52% end-to-end mAP that of uploading all images to the cloud, and misses only about 2.16%-6% of objects while saving half of the bandwidth resources.

## REFERENCES

[1] A. G. Howard, M. Zhu, B. Chen, D. Kalenichenko, W. Wang, T. Weyand, M. Andreetto, and H. Adam, "Mobilenets: Efficient convolutional neural networks for mobile vision applications," 2017.

[2] B. G. Chun, S. Ihm, P. Maniatis, M. Naik, and A. Patti, "Clonecloud: Elastic execution between mobile device and cloud," in *Conference on Computer Systems*, 2011, pp. 301–304.

[3] K. Boos, D. Chu, and E. Cuervo, "Flashback: Immersive virtual reality on mobile devices via rendering memoization," in *the 14th Annual International Conference*, 2016.

[4] F. Liu, S. Peng, J. Hai, L. Ding, Y. Jie, N. Di, and L. Bo, "Gearing resource-poor mobile devices with powerful clouds: architectures, challenges, and applications," *IEEE Wireless Communications*, vol. 20, no. 3, pp. 14–22, 2013.

[5] M. Satyanarayanan, "A brief history of cloud offload: A personal journey from odyssey through cyber foraging to cloudlets," *ACM SIGMOBILE Mobile Computing and Communications Review*, vol. 18, no. 4, pp. 19–23, 2015.

[6] Y. Kang, J. Hauswald, G. Cao, A. Rovinski, and L. Tang, "Neurosurgeon: Collaborative intelligence between the cloud and mobile edge," *ACM SIGOPS Operating Systems Review*, pp. 615–629, 2017.

[7] J. H. Ko, T. Na, M. F. Amir, and S. Mukhopadhyay, "Edge-host partitioning of deep neural networks with feature space encoding for resource-constrained internet-of-things platforms," in *2018 15th IEEE International Conference on Advanced Video and Signal Based Surveillance (AVSS)*, 2018, pp. 1–6.

[8] H. Li, C. Hu, J. Jiang, W. Zhi, and W. Zhu, "Jalad: Joint accuracy- and latency-aware deep structure decoupling for edge-cloud execution," *2018 IEEE 24th International Conference on Parallel and Distributed Systems (ICPADS)*, pp. 671–678, 2018.

[9] W. Shi, Y. Hou, S. Zhou, Z. Niu, Y. Zhang, and L. Geng, "Improving device-edge cooperative inference of deep learning via 2-step pruning," 2019.

[10] L. Lai, N. Suda, and V. Chandra, "Cmsis-nn: Efficient neural network kernels for arm cortex-m cpus," 2018.

[11] N. D. Lane, P. Georgiev, and L. Qendro, "Deepear: Robust smartphone audio sensing in unconstrained acoustic environments using deep learning," in *UbiComp '15 ACM International Joint Conference on Pervasive and Ubiquitous Computing - September 07 - 11, 2015*, 2015, pp. 283–294.

[12] S. Liu, Y. Lin, Z. Zhou, K. Nan, and J. Du, "On-demand deep model compression for mobile devices: A usage-driven model selection framework," in *the 16th Annual International Conference*, 2018.

[13] S. Yao, Y. Zhao, H. Shao, S. Liu, D. Liu, L. Su, and T. Abdelzaher, "Fastdeepiot: Towards understanding and optimizing neural network execution time on mobile and embedded devices," *ACM*, pp. 278–291, 2018.

[14] M. Sandler, A. Howard, M. Zhu, A. Zhmoginov, and L. C. Chen, "Mobilenetv2: Inverted residuals and linear bottlenecks," in *2018 IEEE/CVF Conference on Computer Vision and Pattern Recognition (CVPR)*, 2018.

[15] A. Ashok, P. Steenkiste, and F. Bai, "Vehicular cloud computing through dynamic computation offloading," *Computer Communications*, vol. 120, no. MAY, pp. 125–137, 2018.

[16] E. Li, Z. Zhou, and X. Chen, "Edge intelligence: On-demand deep learning model co-inference with device-edge synergy," 2018.

[17] S. Zhang, Y. Li, X. Liu, S. Guo, and D. Wu, "Towards real-time cooperative deep inference over the cloud and edge end devices," *Proceedings of the ACM on Interactive, Mobile, Wearable and Ubiquitous Technologies*, 2020.

[18] W. Liu, D. Anguelov, D. Erhan, C. Szegedy, S. Reed, C. Y. Fu, and A. C. Berg, "Ssd: Single shot multibox detector," *European Conference on Computer Vision*, 2016.

[19] J. Redmon, S. Divvala, R. Girshick, and A. Farhadi, "You only look once: Unified, real-time object detection," *IEEE*, 2016.

[20] R. Girshick, J. Donahue, T. Darrell, and J. Malik, "Rich feature hierarchies for accurate object detection and semantic segmentation," *IEEE Computer Society*, 2013.

[21] K. Simonyan and A. Zisserman, "Very deep convolutional networks for large-scale image recognition," *Computer Science*, 2014.

[22] "Serving dnns in real time at datacenter scale with project brainwave," *IEEE Micro*, vol. 38, no. 2, pp. 8–20, 2018.

[23] K. Hazelwood, S. Bird, D. Brooks, S. Chintala, and U. Diril, "Applied machine learning at facebook: A datacenter infrastructure perspective," in *IEEE International Symposium on High Performance Computer Architecture*, 2018, pp. 620–629.

[24] O. Jian, S. Lin, Q. Wei, W. Yong, and J. Song, "Sda: Software-defined accelerator for large-scale dnn systems," in *2014 IEEE Hot Chips 26 Symposium (HCS)*, 2016, pp. 1–23.

[25] N. P. Jouppi, C. Young, N. Patil, D. Patterson, and G. A. E. Al, "In-datacenter performance analysis of a tensor processing unit," in *Computer architecture news*, 2017, pp. 1–12.

[26] C. Hu, W. Bao, D. Wang, and F. Liu, "Dynamic adaptive dnn surgery for inference acceleration on the edge," in *IEEE Conference on Computer Communications*, 2019, pp. 1423–1431.

[27] E. Li, L. Zeng, Z. Zhou, and X. Chen, "Edge ai: On-demand accelerating deep neural network inference via edge computing," *IEEE Transactions on Wireless Communications*, vol. 19, no. 1, pp. 447–457, 2020.

[28] G. Hinton, O. Vinyals, and J. Dean, "Distilling the knowledge in a neural network," *Computer Science*, vol. 14, no. 7, pp. 38–39, 2015.

[29] J. Kim, S. Park, and N. Kwak, "Paraphrasing complex network: Network compression via factor transfer," pp. 2760–2763, 2018.

[30] G. Huang, Z. Liu, V. Laurens, and K. Q. Weinberger, "Densely connected convolutional networks," in *IEEE Computer Society*, 2016, pp. 4700–4708.

[31] S. Ren, K. He, R. Girshick, and J. Sun, "Faster r-cnn: Towards real-time object detection with region proposal networks," *IEEE Transactions on Pattern Analysis & amp; Machine Intelligence*, vol. 39, no. 6, pp. 1137–1149, 2017.

[32] H. Law and J. Deng, "Cornernet: Detecting objects as paired keypoints," *International Journal of Computer Vision*, vol. 128, no. 3, pp. 642–656, 2020.

[33] S. Han, J. Pool, J. Tran, and W. J. Dally, "Learning both weights and connections for efficient neural networks," *MIT Press*, pp. 1135–1143, 2015.

[34] T. Y. Lin, P. Goyal, R. Girshick, K. He, and P. Dollár, "Focal loss for dense object detection," *IEEE Transactions on Pattern Analysis amp; Machine Intelligence*, vol. PP, no. 99, pp. 2999–3007, 2017.

[35] T. Forgione, A. Carlier, G, R. Morin, T. O. Wei, V. Charvillat, and P. K.Yadav, "An implementation of a dash client for browsing networked virtual environment," in *2018 ACM Multimedia Conference*, 2018, pp.1263–1264.

[36] V. Vukoti?, C. Raymond, and G. Gravier, "Multimodal and crossmodal representation learning from textual and visual features with bidirectional deep neural networks for video hyperlinking," in *Acm Workshop on Vision amp; Language Integration Meets Multimedia Fusion*, 2016, pp. 37–44.

[37] K. He, X. Zhang, S. Ren, and J. Sun, "Deep residual learning for image recognition," *2016 IEEE Conference on Computer Vision and Pattern Recognition (CVPR)*, pp. 770–778, 2016.

[38] M. Rhu, M. O'Connor, N. Chatterjee, J. Pool, and S. W. Keckler, "Compressing dma engine: Leveraging activation sparsity for training deep neural networks," *IEEE Computer Society*, 2017.

[39] Y. Lecun and L. Bottou, "Gradient-based learning applied to document recognition," *Proceedings of the IEEE*, vol. 86, no. 11, pp. 2278–2324, 1998. recognition challenge," *International Journal of Computer Vision*,

[40] A. Bochkovskiy, C. Y. Wang, and H. Liao, "Yolov4: Optimal speed and accuracy of object detection," 2020.

[41] C. Y. Wang, H. Liao, I. H. Yeh, Y. H. Wu, P. Y. Chen, and J. W. Hsieh, "Cspnet: A new backbone that can enhance learning capability of cnn," 2019.

[42] Kaiming, Zhang, Xiangyu, Shaoqing, and Jian, "Spatial pyramid pooling in deep convolutional networks for visual recognition."

[43] T. Y. Lin, P. Dollar, R. Girshick, K. He, B. Hariharan, and S. Belongie, "Feature pyramid networks for object detection," in *2017 IEEE Conference on Computer Vision and Pattern Recognition (CVPR)*, 2017.

[44] S. Liu, L. Qi, H. Qin, J. Shi, and J. Jia, "Path aggregation network for instance segmentation," *IEEE*, 2018.

[45] S. Laskaridis, S. I. Venieris, M. Almeida, I. LeoNtIadis, and N. D. Lane, "Spinn: Synergistic progressive inference of neural networks over device and cloud," 2020.